\documentclass{article}
\usepackage{spconf,amsmath,graphicx,hyperref}
\usepackage{algorithm}
\usepackage{algorithmic}
\usepackage{amsmath}
\usepackage{amssymb}
\usepackage{booktabs}
\usepackage{multirow}
\usepackage{enumitem}

\title{Hebbian Learning with Global Direction}
\name{
    Wenjia Hua\textsuperscript{1}\textsuperscript{†}, 
    Kejie Zhao\textsuperscript{1}\textsuperscript{†}, 
    Luziwei Leng\textsuperscript{2}, 
    Ran Cheng\textsuperscript{3}, 
    Yuxin Ma\textsuperscript{1}\textsuperscript{*}, 
    Qinghai Guo\textsuperscript{2}\textsuperscript{*}
    \thanks{
    \textsuperscript{†}Wenjia Hua and Kejie Zhao contributed equally. \\
    \hspace*{12pt} \textsuperscript{*}The corresponding authors are Yuxin Ma (mayx@sustech.edu.cn) and Qinghai Guo (guoqinghai@huawei.com). \\
    \hspace*{12pt} The code is available at \url{https://github.com/huawjcn/GHL}.
    }
}
\address{
    \textsuperscript{1}\textit{Department of CSE, Southern University of Science and Technology}, Shenzhen, China\\
    \textsuperscript{2}\textit{ACS Lab, Huawei Technologies Co., Ltd.}, Shenzhen, China\\
    \textsuperscript{3}\textit{Department of Data Science and Artificial Intelligence, Department of Computing,}\\ \textit{Hong Kong Polytechnic University}, Hong Kong, China
}
\begin{document}
\ninept
\maketitle
\begin{abstract}
Backpropagation algorithm has driven the remarkable success of deep neural networks, but its lack of biological plausibility and high computational costs have motivated the ongoing search for alternative training methods. Hebbian learning has attracted considerable interest as a biologically plausible alternative to backpropagation. Nevertheless, its exclusive reliance on local information, without consideration of global task objectives, fundamentally limits its scalability. Inspired by the biological synergy between neuromodulators and local plasticity, we introduce a novel model-agnostic Global-guided Hebbian Learning (GHL) framework, which seamlessly integrates local and global information to scale up across diverse networks and tasks. In specific, the local component employs Oja's rule with competitive learning to ensure stable and effective local updates. Meanwhile, the global component introduces a sign-based signal that guides the direction of local Hebbian plasticity updates. Extensive experiments demonstrate that our method consistently outperforms existing Hebbian approaches. Notably, on large-scale network and complex datasets like ImageNet, our framework achieves the competitive results and significantly narrows the gap with standard backpropagation.
\end{abstract}
\section{Introduction}
\label{sec:intro}

\begin{figure}[t]
    \centering
    \includegraphics[width=\linewidth]{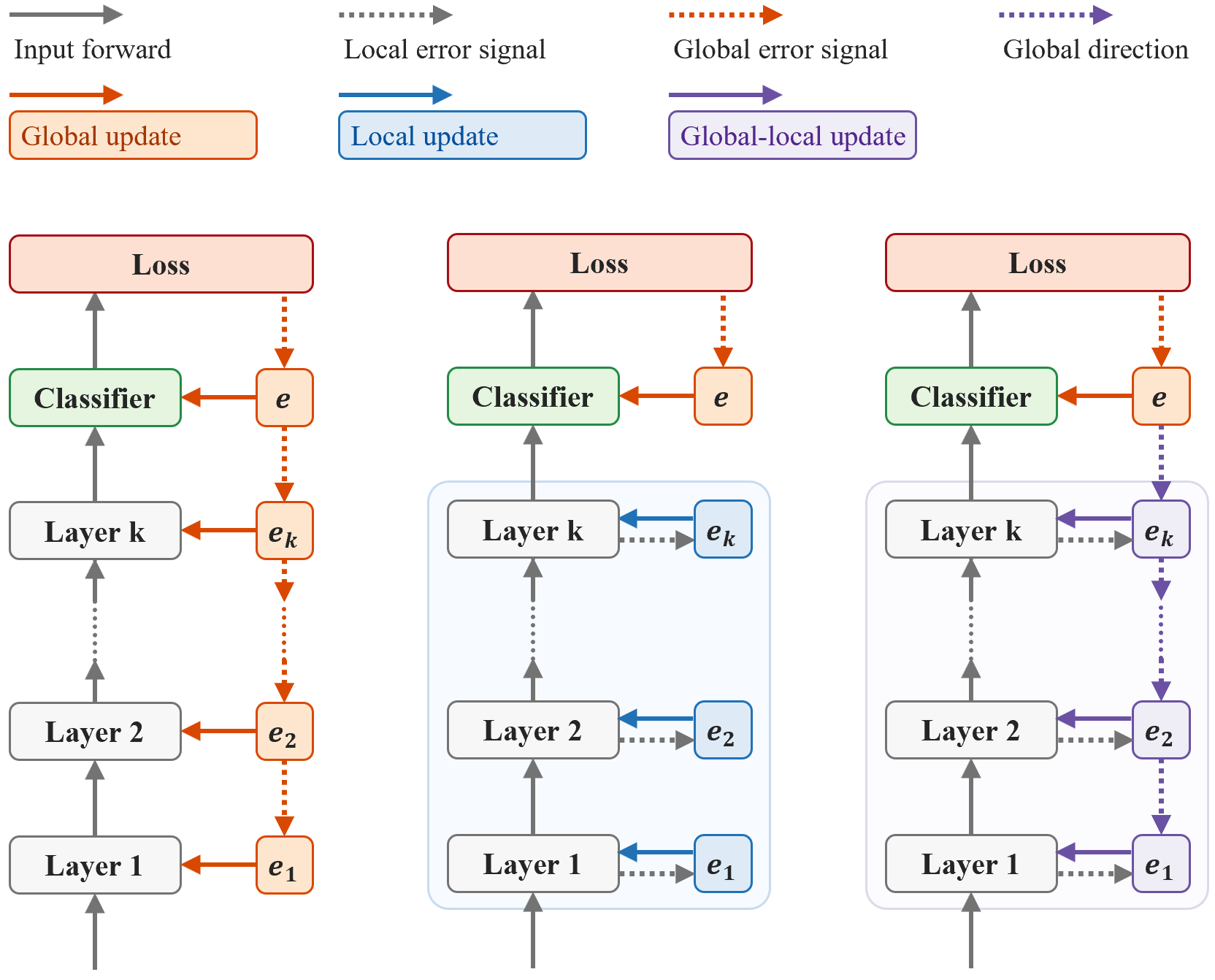}
    \vspace{-12pt}
    \caption{\textbf{Comparison of different training methods.} \textbf{Left:} Standard BP training process, using global error signals to compute precise gradients for weight updates. \textbf{Middle:} Two-factor Hebbian learning approach, which relies only on local information from pre and post synaptic neurons to update weights, without considering global task objectives. \textbf{Right:} Our method, Global-guided Hebbian Learning (GHL), integrates local Hebbian plasticity with a global sign-based modulation.}
    \label{fig:architecture}
\end{figure}

Deep Neural Networks (DNNs) have achieved revolutionary success in recent years, with Backpropagation (BP) playing an important role. Despite its remarkable achievements, BP still has some limitations such as the weight transport problem~\cite{liaoHowImportantWeight2016}. Also, BP's reliance on the precise, global propagation of error signals lacks local plasticity, making it biologically implausible and contrasting sharply with the mechanisms of neural computation in the brain~\cite{whittingtonTheoriesErrorBackPropagation2019, lillicrapBackpropagationBrain2020}. 
To address these limitations, the research community has explored more efficient and biologically inspired training alternatives, such as Feedback Alignment (FA)~\cite{lillicrapRandomSynapticFeedback2016}, Predictive Coding (PC)~\cite{raoPredictiveCodingVisual1999}, and Equilibrium Propagation (EP)~\cite{scellierEquilibriumPropagationBridging2017}. 

Among these approaches, Hebbian learning stands out due to its strong biological plausibility and simple local plasticity rules. 
Early studies have successfully integrated Hebbian learning with deep models like Convolutional Neural Networks (CNNs)~\cite{amatoHebbianLearningMeets2019, laganiComparingPerformanceHebbian2022}. However, these attempts resulted in poor performance. Also, in these early frameworks, the role of Hebbian learning was often restricted, either functioning as an update rule for specific network modules or being used exclusively during a pre-training phase before standard end-to-end training with backpropagation.

To enhance the performance of Hebbian approaches in end-to-end training, researchers have introduced mechanisms such as competitive learning~\cite{songCompetitiveHebbianLearning2000} and anti-Hebbian plasticity~\cite{krotovUnsupervisedLearningCompeting2019}, leading to representative methods including SoftHebb~\cite{moraitisSoftHebbBayesianInference2022, journeHebbianDeepLearning2022}, HWTA-BCM~\cite{nimmoAdvancingBiologicalPlausibility2025}, and FastHebb~\cite{laganiFastHebbScalingHebbian2022, laganiScalableBioinspiredTraining2024}. 
However, like other alternatives to backpropagation, these methods are mostly tested on simple tasks such as MNIST or CIFAR, and with shallow models like MLPs or small CNNs~\cite{guptaBioinspiredLearningBetter2023, illingBiologicallyPlausibleDeep2019}. When facing more complex tasks such as ImageNet or deeper networks, their performance is much worse than backpropagation, showing a critical scalability bottleneck~\cite{journeHebbianDeepLearning2022, qiTrainingDeeperPredictive2025}.

Furthermore, these methods often require specialized network architectures or modules to achieve desirable performance. For instance, DeepSoftHebb~\cite{journeHebbianDeepLearning2022}, which achieves state-of-the-art performance in Hebbian learning, relies heavily on rapidly expanding convolutional channels and specific activation functions. Any modifications to its architecture can lead to a drastic performance drop, sometimes even worse than a single-layer network. This strong dependency on specific network structures severely restricts the practical application and generalization of bio-inspired learning algorithms.

The root of this limitation in existing Hebbian methods lies in their over reliance on local information, which takes the form of two-factor learning~\cite{ororbiaBraininspiredMachineIntelligence2023}, where weight updates are determined solely by the activity of pre and post synaptic neurons. While this pure locality largely circumvents BP's issues like weight transport and update locking, it suffers from a lack of global information. 
From a broader perspective, this is also inconsistent with biological reality. While the brain lacks precise error backpropagation, it does possess feedback mechanisms, such as neuromodulators, that convey global, task-relevant signals~\cite{whittingtonTheoriesErrorBackPropagation2019, lillicrapBackpropagationBrain2020}.

Such a learning rule, which is modulated by a global signal, is also known as the three-factor learning rule. In this rule, synaptic changes depend not only on local activity but also on a third modulatory factor that provides global information. In biological systems, various neuromodulators serve as global signals, controlled by complex feedback mechanisms~\cite{kusmierzLearningThreeFactors2017}. However, in the machine learning community, there is currently no unified implementation for effectively incorporating such global guidance signals into existing artificial neural networks.

To fuse local plasticity with global objectives, we propose a novel and model-agnostic framework called \textbf{G}lobal-guided \textbf{H}ebbian \textbf{L}earning (\textbf{GHL}), designed to bridge the gap between theoretical research and practical machine learning implementation. In our framework, the local update signal is provided by Hebbian learning, using competitive learning and Oja’s rule. Inspired by the biological phenomenon where neuromodulators can `flip' the polarity of synaptic plasticity~\cite{zaitsevInhibitionSlowAfterhyperpolarization2012, brzoskoRetroactiveModulationSpike2015, brzoskoNeuromodulationSpiketimingdependentPlasticity2019}, we instantiate the global guidance signal as a sign signal derived from the global task gradient. This global sign signal provides only directional information (i.e., potentiation or depression) without requiring the precise gradient magnitude.

We extensively benchmarked our proposed framework against state-of-the-art Hebbian methods, demonstrating superior performance across multiple network architectures and achieving accuracy comparable to end-to-end BP. More importantly, our framework demonstrates strong scalability and generalization, and its performance is less affected by changes in network structure. We successfully applied high-performance Hebbian learning to the large and complex ImageNet task, reducing the performance gap with standard backpropagation to within 4\% on ResNet-50. In addition, we are the first to successfully scale a bio-inspired learning algorithm to an extremely deep ResNet-1202 architecture. These results push the boundaries of bio-inspired learning.

Our main contributions are as follows:
\begin{itemize}[nosep, leftmargin=14pt, labelsep=7pt]
    \item We propose a novel, model-agnostic Global-guided Hebbian Learning (GHL) framework, which guides local synaptic plasticity using the sign of the global gradient. This approach naturally combines global optimization objectives with local update rules.
    \item Through extensive experiments, we demonstrate that our framework not only outperforms existing Hebbian learning algorithms but is also scalable to challenging datasets like ImageNet, greatly improving the practical use of bio-inspired learning.
    \item Our work bridges the gap between neuroscience-inspired learning mechanisms and those used in machine learning, offering new possibilities for interpreting learning processes in biological systems.
\end{itemize}

\section{Method}
\label{sec:method}

\subsection{Overview of GHL framework}

The three-factor learning rule provides a model for synaptic plasticity. It posits that the change in a synaptic weight $\Delta w_{ik}$ is governed by two components: a local Hebbian term $H(\text{pre}_i, \text{post}_k)$, which depends on pre- and post-synaptic activities, and a global modulatory signal $G(m)$. The three-factor learning rule is a high-level abstraction of biological neural mechanisms, whose exact components are not clearly defined. Similarly, how to effectively apply this rule in artificial neural networks remains a major open question.

In biological systems, various neuromodulators such as dopamine, noradrenaline, acetylcholine, and serotonin play the role of $G(m)$. Biological evidence has shown that dopamine can act as a `gate', determining whether Hebbian plasticity occurs~\cite{pawlakTimingNotEverything2010}. Furthermore, studies have revealed that dopamine can also `reverse' the polarity of synaptic plasticity~\cite{brzoskoNeuromodulationSpiketimingdependentPlasticity2019}, for instance by converting long-term depression (LTD) into long-term potentiation (LTP) under certain conditions~\cite{brzoskoRetroactiveModulationSpike2015, zhangGainSensitivityLoss2009}. Moreover, adrenergic and cholinergic stimulation can also alter the direction of plasticity~\cite{zaitsevInhibitionSlowAfterhyperpolarization2012}.

Inspired by these biological mechanisms, we abstract this complex modulation, whether it be gating or polarity reversal, into a concise form: a sign. Therefore, in our framework, we instantiate the modulatory factor $G(m)$ as $\text{sign}(m)$. Thus, our Global-guided Hebbian Learning (GHL) method is formulated as:
\begin{align}
\Delta w_{ik} &= \eta \cdot \text{sign}(m) \cdot H(\text{pre}_i,\text{post}_k),
\end{align}
where $\eta$ is the learning rate. We integrate the GHL method into an end-to-end training framework. The overall training architecture is illustrated in Figure~\ref{fig:architecture}.

\subsection{Competitive Hebbian learning}

Our local update is grounded in the principles of competitive Hebbian learning.
The core tenet of Hebbian learning, `neurons that fire together, wire together,' is fundamentally expressed as $\Delta w_{ik} \propto y_k \cdot x_i$, where $x_i$ is the presynaptic activation and $y_k$ is the postsynaptic activation. 
However, this primitive form is inherently unstable, as the weights tend to either grow unbounded or decay to zero. To counteract this, we incorporate Oja's rule:
\begin{align}
    \Delta w_{ik}^{\text{Oja}} &= \eta \cdot y_k \cdot x_i - \eta \cdot y_k^2 \cdot w_{ik} \\
    &= \eta \cdot y_k \cdot (x_i - y_k \cdot w_{ik}),
\end{align}
where the second term $\eta \cdot y_k^2 \cdot w_{ik}$ acts as a normalization or `forgetting' term that stabilizes the learning process, preventing the weights from growing excessively.

To further enable neurons to specialize and learn distinct input patterns, we incorporate a competitive learning mechanism. This mimics the biological phenomena of lateral inhibition and Winner-Take-All (WTA). We adopt a smoothed variant, Soft Winner-Take-All (SWTA), which enforces competition among neurons in the same layer using the softmax function to produce the competitive output $u_k = \text{Softmax}(y_k)$.

By integrating the competitive mechanism with Oja's rule, we arrive at our final local update rule:
\begin{align}
    \Delta w_{ik}^{\text{SWTA}} &= \eta \cdot u_k \cdot (x_i - y_k \cdot w_{ik}). \label{eq:swta_update}
\end{align}

This rule ensures that only the `winning' neurons (those with large $u_k$ values) significantly update their weights, thereby achieving feature selectivity and stable learning.

\begin{figure}[t]
    \vspace{-9pt}
    \begin{algorithm}[H]
        \caption{GHL Training}
        \label{alg:algorithm}
        \textbf{Hyperparameters:} \\
        \hspace*{10pt} Learning rate $\eta$, Softmax temperature $\tau$ \\
        \textbf{Input:} \\
        \hspace*{10pt} Training dataset $D$, Network weights $W_{old}$ \\
        \textbf{Output:} \\
        \hspace*{10pt} Network weights $W_{new}$  
        \begin{algorithmic}[1]
            \FOR{each training batch $(X, Y)$ in $D$}
                \STATE \textbf{Forward Pass:} \\
                       \hspace*{15pt} For each output neuron $k$,
                \STATE \hspace*{15pt} $y_k = \sum_i w_{ik} x_i$. \hspace*{89.5pt} \COMMENT {activation}
                \STATE \textbf{Hebbian Update:} \\
                       \hspace*{15pt} For each output neuron $k$,
                \STATE \hspace*{15pt} $u_k = \frac{e^{y_k / \tau}}{\sum_{l=1}^{K} e^{y_l / \tau}}$. \hspace*{52pt} \COMMENT {competition output} \\
                       \hspace*{15pt} For each synaptic weight $w_{ik}$,
                \STATE \hspace*{15pt} $\Delta w_{ik}^{(\text{Hebb})} = u_k \cdot (x_i - y_k \cdot w_{ik})$. \hspace*{20pt} \COMMENT {local update}
                \STATE \textbf{Global Update:} \\
                       \hspace*{15pt} For network output $\hat{Y}$,
                \STATE \hspace*{15pt} $L = \text{Loss}(\hat{Y}, Y)$. \hspace*{80pt} \COMMENT {Global loss}
                \STATE \hspace*{15pt} $G = \frac{\partial L}{\partial W}$. \hspace*{92pt} \COMMENT {Global gradient}
                \STATE \textbf{Sign Modulation:}
                \STATE \hspace*{15pt} $M = \text{sign}(G)$. \hspace*{72pt} \COMMENT {Global direction}
                \STATE \hspace*{15pt} $\Delta W^{(\text{GHL})} = M \odot \left| \Delta W^{(\text{Hebb})} \right|$.
                \STATE \textbf{Final Update:}
                \STATE \hspace*{15pt} $W = W + \eta \cdot \Delta W^{(\text{GHL})}$.
            \ENDFOR
        \end{algorithmic}
    \end{algorithm}
    \vspace{-16pt}
\end{figure}

\subsection{Global direction as a third factor}

Inspired by biological systems, where various neuromodulators can reverse the polarity of local synaptic plasticity, we conceptualize global guidance $m$ in the GHL framework as sign-based information. In biological neural networks, substances such as dopamine, noradrenaline, and acetylcholine, serve this modulatory function, but they are generated and transmitted through complex biological mechanisms. As no direct analog for such feedback currently exists in artificial neural networks, we employ the sign of the backpropagated gradient as an effective proxy for this global signal.

In the end-to-end training of neural networks, the global error stems from the discrepancy between network outputs and ground-truth labels, quantified by a loss function $L(y, \hat{y})$. Accordingly, the global guidance $m$ in our framework is derived from the gradient of this global loss, $\nabla_w L$. 
The final weight update rule for a connection from a presynaptic neuron $i$ to a postsynaptic neuron $k$ can be precisely formulated as:
\vspace{-8pt}
\begin{align}
    \Delta w_{ik} &= \eta \cdot \text{sign}(m) \cdot \left| \Delta w_{ik}^{\text{SWTA}} \right| \\
    &= \eta \cdot \text{sign} \left( \frac{\partial L}{\partial w_{ik}} \right) \cdot \left| u_k \cdot (x_i - y_k \cdot w_{ik}) \right|,
\end{align}
\vspace{-8pt}

where $\Delta w_{ik}^{\text{SWTA}}$ is the local Hebbian update which provides the update's magnitude, modulated by the sign of the global gradient signal $\partial L/\partial w_{ik}$.

In our GHL framework, the primary update is computed by the local Hebbian rule, while the sign-based gradient signal serves only as a modulatory factor, indicating whether synaptic changes should be potentiated (positive) or depressed (negative).
The use of the gradient's sign shares similarities with SignSGD~\cite{bernsteinSignSGDCompressedOptimisation2018a, bernsteinSignSGDMajorityVote2019}, which is primarily designed to reduce communication overhead. However, GHL differs fundamentally in terms of update magnitude. Specifically, GHL uses local Hebbian rules to dynamically adjust update magnitudes, avoiding the instability of fixed-step updates. While SignSGD often relies on high-precision internal computations and error compensation mechanisms to recover the original gradient information, GHL only uses the directional information (i.e., the sign) from the gradient, without requiring its precise magnitude.

The final network training procedure is detailed in Algorithm~\ref{alg:algorithm}. It should be noted that, owing to the locality of Hebbian updates, they can be computed at any stage during the forward and backward pass, and the approach shown in the paper is one such reference implementation.

\section{Experiments and Results}
\label{sec:experiment}

\subsection{Comparison with existing Hebbian methods on CIFAR-10 and CIFAR-100 datasets}

To evaluate our method, we first conducted a rigorous comparison against state-of-the-art Hebbian algorithms on the CIFAR-10/100 datasets~\cite{krizhevskyLearningMultipleLayers2009}. These methods include SoftHebb~\cite{moraitisSoftHebbBayesianInference2022, journeHebbianDeepLearning2022}, FastHebb (SWTA-FH/HPCA-FH)~\cite{amatoHebbianLearningMeets2019, laganiFastHebbScalingHebbian2022, laganiScalableBioinspiredTraining2024}, and HWTA-BCM~\cite{nimmoAdvancingBiologicalPlausibility2025}. To ensure a fair comparison, we replicated two primary baseline network architectures from the literature:\\
\textbf{DeepHebb}~\cite{journeHebbianDeepLearning2022} with 3 convolutional layers (96, 384, 1536 channels) and a linear classifier.\\
\textbf{FastHebb}~\cite{laganiScalableBioinspiredTraining2024} with 4 convolutional layers (96, 128, 192, 256 channels), a fully-connected layer (4096 neurons) and a linear classifier.

As shown in Table \ref{tab:Result_CIFAR10_100_Main}, on both baseline networks, GHL demonstrates a significant accuracy advantage over all other Hebbian competitors and achieves a performance level comparable to that of standard end-to-end backpropagation (E2E BP).

\begin{table}[ht]
    \vspace{-3pt}
    \centering
    \begin{tabular}{c | c r@{\hskip 1pt}l r@{\hskip 1pt}l}
        \toprule
        \parbox{40pt}{\centering Arch} & \parbox{50pt}{\centering Method} & \multicolumn{2}{c}{CIFAR-10} & \multicolumn{2}{c}{CIFAR-100} \\ 
        \midrule
        \multirow{3}{*}{\parbox{40pt}{\centering DeepHebb \\~\cite{moraitisSoftHebbBayesianInference2022}}} 
        & SoftHebb      & 80.36 & (-6.62) & 55.94 & (-7.64) \\
        & \textbf{GHL(Ours)}     & 86.41 & \textbf{(-0.57)} & 62.40 & (-1.18) \\
        & E2E BP        & 86.98 &   & 63.58 &  \\
        \midrule
        \multirow{4}{*}{\parbox{40pt}{\centering FastHebb\\~\cite{laganiScalableBioinspiredTraining2024}}} 
        & SWTA-FH       & 55.23 &  (-29.48) & 21.04 & (-36.45) \\
        & HPCA-FH       & 67.73 &  (-16.98) & 37.47 & (-20.02) \\
        & \textbf{GHL(Ours)}     & 83.96 &  (-0.75) & 57.23 & \textbf{(-0.26)} \\
        & E2E BP        & 84.71 &  & 57.49 &  \\       
        \bottomrule
    \end{tabular}
    \vspace{-4pt}
    \caption{CIFAR-10/100 test accuracy (\%) and difference from standard backpropagation.}
    \vspace{-3pt}
    \label{tab:Result_CIFAR10_100_Main}
\end{table}

To further investigate the generality of our framework, we applied our method to the VGG-16 and ResNet-20 networks on CIFAR-10. Notably, this comparison against other Hebbian-based algorithms does not enforce a unified architecture, aiming instead to benchmark each method under its reported optimal settings. As shown in Table~\ref{tab:Result_CIFAR10_Comparison}, our method surpasses the highest accuracies reported by these baselines.

\begin{table}[ht]
    \vspace{-3pt}
    \centering
    \begin{tabular}{l l@{\hskip 5pt}c c}
        \toprule
        Method                                                                              & Arch                  & Layers & Acc \\
        \midrule
        \textbf{GHL(Ours)}                                                                  & VGG-16                & 16  & \textbf{89.48} \\
        GHL(Ours)                                                                           & ResNet-20             & 20  & 86.72 \\
        \midrule
        HaH~\cite{cekicNeuroInspiredDeepNeural2022, cekicRobustInterpretableNeural2022}     & VGG-16                & 16  & 87.3    \\
        HWTA-BCM~\cite{nimmoAdvancingBiologicalPlausibility2025}                            & DeepHebb              & 4   & 76.0    \\
        IB~\cite{daruwallaInformationBottleneckbasedHebbian2024}                            & CNN (128/256 channel) & 3   & 61.0    \\
        \bottomrule
    \end{tabular}
    \vspace{-4pt}
    \caption{CIFAR-10 test accuracy (\%) and broad comparison with other Hebbian-based methods.}
    \vspace{-6pt}
    \label{tab:Result_CIFAR10_Comparison}
\end{table}

\subsection{Scalability on complex datasets and deeper networks}

To evaluate the scalability of our framework, we first conducted experiments on the large-scale ImageNet (ILSVRC 2012) dataset~\cite{dengImageNetLargescaleHierarchical2009a}, which is significantly more challenging than the CIFAR datasets. Previous Hebbian-based methods have struggled on this benchmark. As shown in Table~\ref{tab:Result_ImageNet_Main}, our method outperforms existing Hebbian variants and substantially narrows the gap with standard BP.

\begin{table}[ht]
    \vspace{-3pt}
    \centering
    \begin{tabular}{c | c r@{\hskip 1pt}l r@{\hskip 1pt}l }
        \toprule
        \parbox{42pt}{\centering Arch } & \parbox{50pt}{\centering Method} & \multicolumn{2}{c}{Top-1 Acc} & \multicolumn{2}{c}{Top-5 Acc} \\
        \midrule
        \multirow{3}{*}{\parbox{42pt}{\centering DeepHebb\\~\cite{journeHebbianDeepLearning2022}}} 
        & SoftHebb      & 26.08 & (-32.76)& 41.05 & (-39.43) \\
        & GHL(Ours)     & 53.71 & (-5.13) & 76.02 & (-4.46) \\
        & BP E2E        & 58.84 &         & 80.48 &  \\
        \midrule
        \multirow{4}{*}{\parbox{42pt}{\centering FastHebb\\~\cite{laganiScalableBioinspiredTraining2024}}} 
        & SWTA-FH       &  4.75 & (-51.89) & 12.49 & (-67.15) \\
        & HPCA-FH       & 19.42 & (-37.22) & 42.22 & (-37.42) \\
        & GHL(ours)     & 50.80 & (-5.84)  & 74.56 & (-5.08) \\
        & BP E2E        & 56.64 &          & 79.64 &  \\
        \midrule
        \multirow{2}{*}{\parbox{42pt}{\centering VGG-16\\~\cite{simonyanVeryDeepConvolutional2015}}}
        & GHL(ours)     & 65.58 & (-5.92)  & 86.49 & (-3.61) \\
        & BP E2E        & 71.50 &          & 90.10 &  \\
        \midrule
        \multirow{2}{*}{\parbox{42pt}{\centering ResNet-18\\~\cite{heDeepResidualLearning2016}}} 
        & GHL(ours)     & 65.78 & (-3.79) & 86.45 & (-2.79) \\
        & BP E2E        & 69.57 &         & 89.24 &  \\
        \midrule
        \multirow{2}{*}{\parbox{42pt}{\centering ResNet-50\\~\cite{heDeepResidualLearning2016}}} 
        & \textbf{GHL(Ours)}    & 73.14 & \textbf{(-2.85)} & 91.04 & \textbf{(-1.94)} \\
        & BP E2E                & 75.99 &                  & 92.98 & \\
        \bottomrule
    \end{tabular}
    \vspace{-4pt}
    \caption{ImageNet (ILSVRC 2012) Top-1/5 validation accuracy (\%) and difference from standard backpropagation.}
    \vspace{-3pt}
    \label{tab:Result_ImageNet_Main}
\end{table}

For depth scalability, we followed the original ResNet paper~\cite{heDeepResidualLearning2016}, testing depths ranging from 20 to 1202 layers on CIFAR-10. Table~\ref{tab:CIFAR10_Depths} shows that our method maintains robust performance without significant degradation as depth increases, even with extremely deep networks, confirming the scalability and effectiveness of our GHL framework.

\begin{table}[ht]
    \vspace{-3pt}
    \centering
    \setlength{\tabcolsep}{4pt}
    \begin{tabular}{l|cc|c@{\hskip 5pt}c@{\hskip 5pt}c@{\hskip 5pt}c@{\hskip 5pt}c@{\hskip 5pt}c@{\hskip 3pt}}
        \toprule
        Arch & \multicolumn{2}{c|}{VGG~\cite{simonyanVeryDeepConvolutional2015}} & \multicolumn{6}{c}{ResNet~\cite{heDeepResidualLearning2016}} \\
        Layers  & 14 & 16 & 20 & 32 & 44 & 56 & 110 & 1202 \\
        \midrule
        Params  & 14.71 & 33.63 & 0.27 & 0.46 & 0.66 & 0.85 & 1.72 & 19.33 \\
        Acc     & 89.29 & \textbf{89.48} & 86.72 & 86.86 & 87.09 & 87.17 & 87.21 & 86.97 \\
        \bottomrule
    \end{tabular}
    \vspace{-4pt}
    \caption{CIFAR-10 test accuracy (\%) on networks of varying depths and parameters (M). The VGG-14 network is a variant of VGG-16, with the last two fully connected layers removed.}
    \vspace{-10pt}
    \label{tab:CIFAR10_Depths}
\end{table}

\subsection{The architecture’s impact}

\begin{figure}[t]
    \centering
    \includegraphics[width=\linewidth]{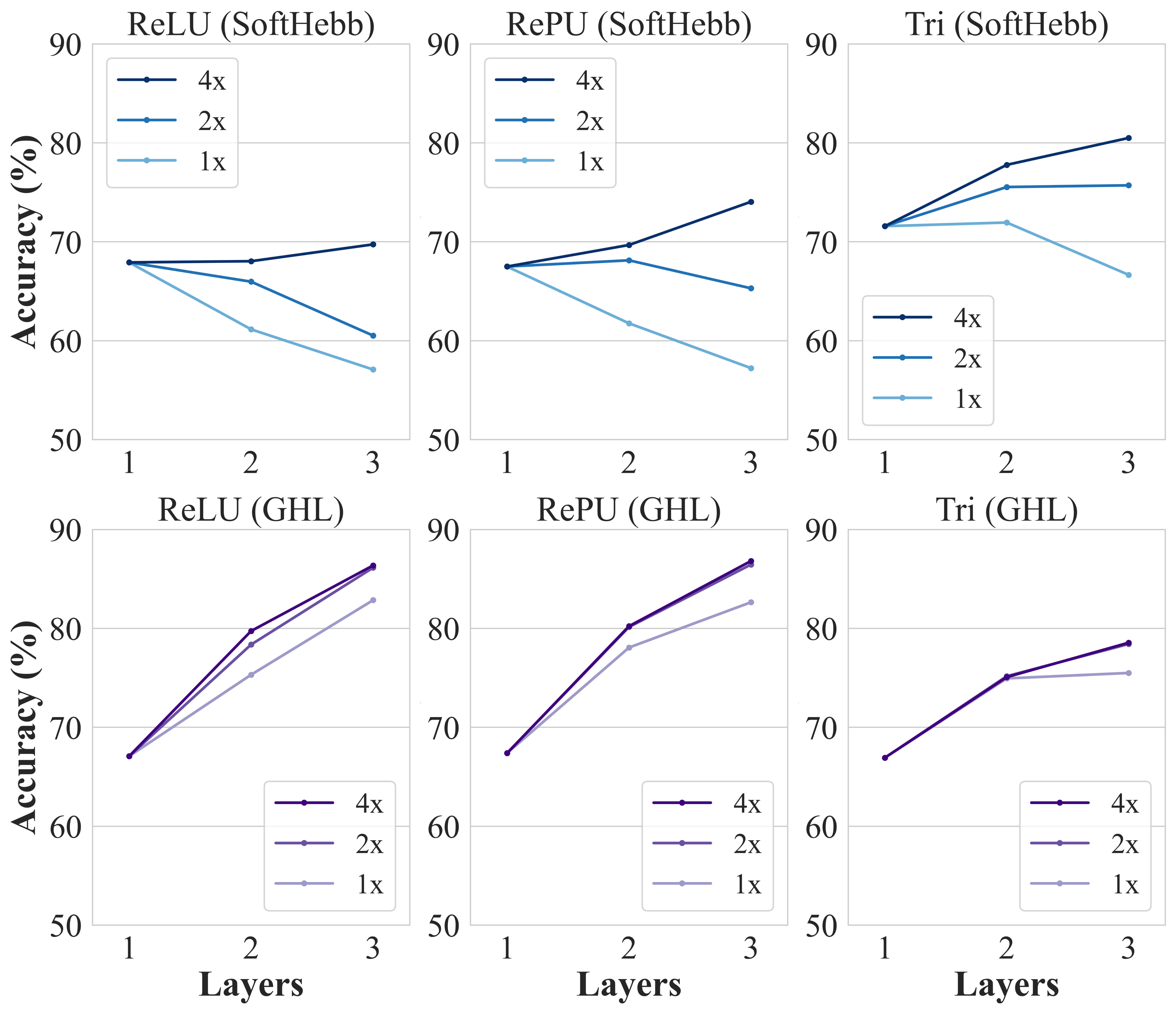}
    \vspace{-20pt}
    \caption{\textbf{Performance comparison of different architectures using SoftHebb and GHL.} The x-axis represents network depth (the number of convolutional layers), and the y-axis shows Top-1 accuracy. The architectures tested have initial channel counts of 96, which are then expanded by factors of 1, 2, and 4, respectively.}
    \vspace{-3pt}
    \label{fig:Aablation_Architecture}
\end{figure}

In our experiments, we adopted architectures from existing literature for a fair comparison. Most of these architectures are specifically designed, as Hebbian methods are often sensitive to design choices such as channel configuration and activation functions. For example, DeepHebb~\cite{journeHebbianDeepLearning2022} achieves strong results on CIFAR-10 by using a shallow network with rapidly increasing channel widths, and a custom `Triangle' activation function defined as:
\begin{align*}
    \text{RePU}(u_j) &= \max(0, u_j^p) \\
    \text{Triangle}(u_j) &= \text{RePU}(u_j - \bar{u})
\end{align*}
where $u_j$ is the activation of the $j$-th channel and $\bar{u}$ is the mean activation across channels. We conducted ablation studies with various activations and architectures. As shown in Figure~\ref{fig:Aablation_Architecture}, while SoftHebb only improves with depth in certain cases, our GHL framework consistently achieves performance gains across all tested settings.

\subsection{Ablation on local information}

Previous experiments have shown that our GHL framework, which incorporates global directional guidance, significantly outperforms purely Hebbian updates that use only local signals. To further isolate the contribution of each component, we conducted an ablation study where the network was updated using only the global signal. This setup is similar to SignSGD, but with an important difference: for a fair comparison and to match the GHL framework's design, we used only the sign information (+1, -1) in our updates, without gradient norms, momentum, or error compensation mechanisms commonly found in SignSGD-like methods. 

As shown in Table~\ref{tab:Ablation_Sign}, using the sign-only update results in lower performance compared to the full GHL framework. Notably, this performance gap widens as the network's scale increases.

\begin{table}[ht]
    \vspace{-3pt}
    \centering
    \begin{tabular}{l|c@{\hskip 8pt}c|c@{\hskip 8pt}c@{\hskip 8pt}c}
        \toprule
        \multirow{2}{*}{Method} & \multicolumn{2}{c|}{Signal} & \multicolumn{3}{c}{Arch} \\
                                & global & local              & DeepHebb & FastHebb & ResNet-20 \\
        \midrule
        GHL    & \checkmark & \checkmark & 86.41    & 83.96    & \textbf{86.72}    \\
        Sign   & \checkmark &            & 81.01    & 73.39    & 63.22    \\
        \bottomrule
    \end{tabular}
    \vspace{-4pt}
    \caption{CIFAR-10 test accuracy (\%) from ablation study on the impact of global and local signals.}
    \vspace{-9pt}
    \label{tab:Ablation_Sign}
\end{table}

\section{Discussion}

In this work, we presented the Global-guided Hebbian Learning (GHL) framework to address the scalability and generalization limitations of existing Hebbian methods. This framework provides new insights into biological learning processes and offers a viable approach for learning on neuromorphic hardware. Future directions include extending GHL to modern network architectures like Transformers and exploring other local and global learning mechanisms.

\bibliographystyle{IEEEbib}
\bibliography{refs}

\end{document}